# Introduction to Arabic Speech Recognition
## Using CMUSphinx System

H. Satori [1, 2], M. Harti [1, 2], and N. Chenfour [2].

*(1) : UFR Informatique et Nouvelles Technologies d'Information et de Communication
B.P. 1796, Dhar Mehraz Fès Morocco.*

*(2) : Département de Mathématiques et Informatique, Faculté des Sciences, B.P. 1796, Dhar Mehraz Fès, Morocco*
E-mail: hsnsatori@yahoo.fr

*Abstract*— **In this paper Arabic was investigated from the speech recognition problem point of view. We propose a novel approach to build an Arabic Automated Speech Recognition System (ASR). This system is based on the open source CMU Sphinx-4, from the Carnegie Mellon University. CMU Sphinx is a large-vocabulary; speaker-independent, continuous speech recognition system based on discrete Hidden Markov Models (HMMs). We build a model using utilities from the OpenSource CMU Sphinx. We will demonstrate the possible adaptability of this system to Arabic voice recognition.**

## I. INTRODUCTION

Automatic Speech Recognition (ASR) is a technology that allows a computer to identify the words that a person speaks into a microphone or telephone. It has a wide area of applications: Command recognition (Voice user interface with the computer), Dictation, Interactive Voice Response, it can be used to learn a foreign language. ASR can help also, handicapped people to interact with society. It is a technology which makes life easier and very promising [1].

View the importance of ASR too many systems are developed, the most popular are: Dragon Naturally Speaking, IBM Via voice, Microsoft SAPI. Open source speech recognition systems are available too, such as HTK [2], ISIP [3], AVCSR [4] and CMU Sphinx-4 [5-7]. We are interested to this last, which is based on Hidden Markov Models (HMMs) [8]. A Hidden Markov Model (HMM) is a statistical model where the system being modeled is assumed to be a Markov process with unknown parameters, and the challenge is to determine the hidden parameters, from the observable parameters, based on this assumption. The extracted model parameters can then be used to perform further analysis, for example for pattern recognition applications. Its extension into foreign languages (English is the standard) represent a real research challenge area.

Although Arabic is currently one of the most widely spoken language in the world, there has been relatively little speech recognition research on Arabic compared to the other languages [9-11]. The first works on Arabic ASR has concentrated on developing recognizers for modern standard Arabic (MSA). The most difficult problems in developing highly accurate ASRs for Arabic are the predominance of non diacritized text material, the enormous dialectal variety, and the morphological complexity.

D. Vergyri et al. investigate the use of morphology-based language model at different stages in a speech recognition system for conversational Arabic [9]. K. Kirchhoff et al. [10] investigate the recognition of dialectal Arabic and study the discrepancies between dialectal and formal Arabic in the speech recognition point of view. D. Vergyri et al [11] investigate the automatic diacritizing Arabic text for use in acoustic model training for ASR.

CMU (Carnegie Mellon University) Sphinx speech recognition system is freely available and currently is one of the most robust speech recognizers in English. This system enables research groups with modest budgets to quickly begin conducting research and developing applications. In this work we attempt to build an ASR based on CMU Sphinx for the Arabic language. We describe our experience for the extension of this system into Arabic language.

## II. PHINX

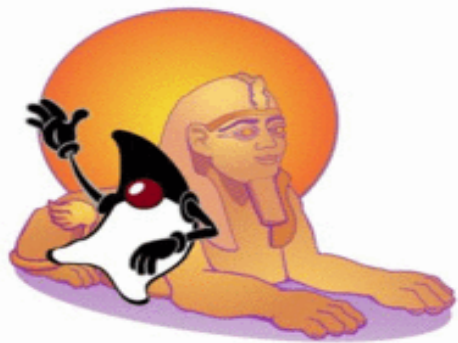

**Fig. 1:** CMUSphinx logo.

**Fig. 2:** Sphinx-4 Architecture, the main blocks are the FrontEnd, the Decoder, and the Linguist.

*A. Presentation*

*1) Sphinx-4*

Sphinx-4 speech recognition system has been jointly developed by the Carnegie Mellon University, Sun Microsystems Laboratories and Mitsubishi Electric Research Laboratories (M.E.R.L). It has been built entirely in Java TM programming language. Since his starting, the Sphinx Group is dedicated to release a set of reasonably mature, speech components. Those provide a basic technology level to anyone interested in creating speech recognition systems. Since 2000, first with CMU Sphinx I, CMU Sphinx II, CMU SphinxTrain then CMU Sphinx III and CMUSphinx-4, a large part of CMU Sphinx project has been made available as open source packages [12-14].

*2) SphinxTrain*

SphinxTrain is the acoustic training environment for CMU Sphinx (for sphinx2, sphinx3 and sphinx4), which was first publicly released on June 7th, 2001. It is a suite of programs, script and documentation for building acoustic models from data for the Sphinx suite of recognition engines. With this contribution, people should be able to build models for any language and condition for which there is enough acoustic data.

It is not possible to proceed to the recognition without having an acoustic model, which is necessary to compare the data coming from FrontEnd (see Fig. 2). This model should be prepared using Sphinx Train tool.

*B. Architecture*

The Sphinx-4 architecture has been designed with a high degree of flexibility and modularity. Each labelled element in Figure 2 represents a module that can be easily replaced, allowing researchers to experiment with different module implementations without needing to modify other portions of the system. The main blocks in Sphinx-4 architecture are frontend, decoder and Linguist.

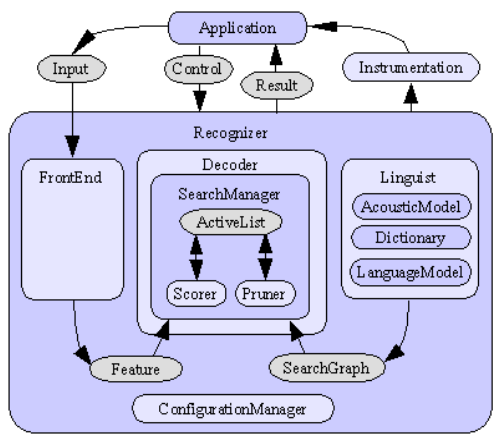

**Front End:** it parameterizes an impute signal (e.g. audio) into a sequence of output features. It performs Digital Signal Processing (DSP) on the incoming data.

--*Feature:* The outputs of the front end are features, used for decoding in the rest of the system.

**Linguist:** Or knowledge base, it provides the information the decoder needs to do its job. It is made up of three modules which are:

--*Acoustic Model*: Contains a representation (often statistical) of a sound, created by training using many acoustic data.

--*Dictionary:* It is responsible for determining how a word is pronounced.

--*Language Model:* It contains a representation (often statistical) of the probability of occurrence of words.

**Search Graph:** The graph structure produced by the linguist according to certain criteria (e.g., the grammar), using knowledge from the dictionary, the acoustic model, and the language model.

**Decoder:** It is the main bloc of the Sphinx-4 system, which performs the bulk of the work. It reads features from the front end, couples this with data from the knowledge base and feedback from the application, and performs a search to determine the most likely sequences of words that could be represented by a series of features.

*C. Installation*

*1) Sphinx-4*

Sphinx-4 can be downloaded either in binary format or in source codes [15] It was compiled and tested on several versions of Linux and on Windows operating systems. Running, building and testing sphinx-4 requires additional software:

- Java 2 SDK, Standard Edition 5.0 [16].
- Java Runtime Environment (JRE)
- Ant: the tool to facilitate compilation and the implementation of sphinx-4 system [17].

*2) SphinxTrain*

We can download the SphinxTrain CMU training package freely [13]. The execution of SphinxTrain requires additional software too:

- Active Perl: To edit scripts provided by Sphinxtrain [18].
- Microsoft Visual Studio: To compile files in Visual C++.

*D. Implementation*

The Sphinx-4 implementation consists of:

- New project from File menu
- Insert the APIs Sphinx-4 into the new project.
- …
- The last step consists of writing a Java code in order to manage and determine components to be used in the system.

### III. APPLICATION FOR ARABIC VOICE RECOGNITION

*A. Arabic language*

Arabic is a Semitic language, and it is one of the oldest languages in the world. It is the fifth widely used language nowadays [19].
Standard Arabic has 34 basic phonemes[1], of which six are vowels, and 28 are consonants [20]. Arabic has fewer vowels than English. It has three long and three short vowels, while American English has at least 12 vowels [21]. Arabic phonemes contain two distinctive classes, which are named pharyngeal and emphatic phonemes. These two classes can be found only in Semitic languages like Hebrew [20-22]. The allowed syllables in Arabic language are: CV, CVC, and CVCC where V indicates a (long or short) vowel while C indicates a consonant. Arabic utterances can only start with a consonant [20].
All Arabic syllables must contain at least one vowel. Also Arabic vowels cannot be initials and can occur either between two consonants or final in a word. Arabic syllables can be classified as short or long. The CV type is a short one while all others are long. Syllables can also be classified as open or closed.
An open syllable ends with a vowel, while a closed syllable ends with a consonant. For Arabic, a vowel always forms a syllable nucleus, and there are as many syllables in a word as vowels in it [23].

*B. Hello Arabic Digit application*

| Digit | Arabic Writing | Pronunciation | Syllables | IPA Representation | No. of Syllables |
|---|---|---|---|---|---|
| 1 | واحد | wā-ḥid | CV-CVC | waː-ḥid | 2 |
| 2 | اثنين | ʔith-nāyn | CVC-CVCC | ʔiθ-niːn | 2 |
| 3 | ثلاثة | thā-lā-thāh | CV-CV-CVC | θa-la-θah | 3 |
| 4 | أربعة | 'aăr-bă-ʿăh | CVC-CV-CVC | ʔar-ba-ʿah | 3 |
| 5 | خمسة | khām-sāh | CVC-CVC | xam-sah | 2 |
| 6 | ستة | sĕt-tăh | CVC-CVC | sit-tah | 2 |
| 7 | سبعة | sŭb-ʿăh | CVC-CVC | sab-ʿah | 2 |
| 8 | ثمانية | thā-mā-nĕ-yah | CV-CV-CV-CVC | θa-ma-ni-jah | 4 |
| 9 | تسعة | tĕs-ăh | CVC-CVC | tis-ʿah | 2 |
| 0 | صفر | şĕfr | CVCC | şifr | 1 |

**Table 1:** Ten Arabic digits, how to pronounce them, type of syllable, IPA representation [24] and number of syllables in every spoken digit [25].

---

[1] A phoneme is the smallest element of speech that indicates a difference in meaning, word, or sentence.

In this paper we describe our experience to create and develop an Arabic Version of CMU Sphinx-4 speech recognition system. In what follows we present a Hello_Arabic_Digit application for the recognition of the ten Arabic digits (Table. 1).

An automatic Speech recognizer system like Sphinx-4 uses three types of language-dependent models:

-- An acoustic model, which represents statistically a range of possible audio representations for the phonemes.
-- A pronunciation dictionary specifying how each word is pronounced in terms of the phonemes in the acoustic model.
-- A language model or grammar model which models patterns of word usage. This is normally customized for the application. Every word in the language model must be in the pronunciation dictionary.

In Hello_Arabic_Digits we proceed to the modification of those three elements to fit to our application.

*1) Corpus preparation*

An in-house corpus was created from all 10 Arabic digits. A number of 6 Moroccan speakers (6 males) were asked to utter all digits 5 times.
Hence, the corpus consists of 5 repetitions of every digit produced by each speaker. Depending on this, the corpus consists of 300 tokens. During the recording session, each utterance was played back to ensure that the entire digit was included in the recorded signal. All the 300 (10 digits · 5 repetitions · 6 speakers) tokens were used for training phases.

*2) Results*

In order to evaluate the performances of the application, we performed some experiments on different individuals (three men) each one of them was asked to utter 10 Arabic digits. We recorded the number of words that were correctly recognized, and then a mean recognition ratio for each tester was calculated (see Table 2).

|    | Test 1 | Test 2 | Test 3 | Mean Recognition Ratio |
|----|--------|--------|--------|------------------------|
| H1 | 9      | 8      | 9      | 86,66%                 |
| H2 | 8      | 9      | 9      | 86,66%                 |
| H3 | 8      | 8      | 9      | 83,33%                 |

**Table 2:** Results of the test of Hello_Arabic_Digit application.

**Fig. 3:** Execution of the Hello_Arabic_Digit application.

Results are very satisfactory taken into account the very small size of the corpus of training (personal corpus) which was used if compared with corpora used for English.

IV. CONCLUSION

To conclude, a spoken Arabic recognition system was designed to investigate the process of automatic speech recognition. This system was based on CMUSphinx-4 from Carnegie Mellon University. An application, *Hello_Arabic_Digit,* was presented to demonstrate the possible adaptability of this system to Arabic speech. We project to extend to application for wide Arabic language recognition, especially for the Moroccan dialect language.